\documentclass[letterpaper, 10 pt, journal, twoside]{IEEEtran}
\usepackage{amsmath} 
\usepackage{amssymb}  
\usepackage{amsfonts}
\usepackage{mathtools}
\usepackage[noadjust]{cite}
\usepackage{booktabs,multirow}
\usepackage{gensymb}
\usepackage{array}
\usepackage{dblfloatfix}
\usepackage{makecell}
\usepackage{hyperref}

\begin{document}

\title{Do You Know the Way? Human-in-the-Loop Understanding for Fast Traversability Estimation in Mobile Robotics}
\author{Andre Schreiber and Katherine Driggs-Campbell%
\thanks{Manuscript received: December 17, 2024; Revised March 14, 2025; Accepted April 8, 2025.}
\thanks{This paper was recommended for publication by Editor Pascal Vasseur upon evaluation of the Associate Editor and Reviewers' comments.
} 
\thanks{Andre Schreiber and Katherine Driggs-Campbell are with the Department of Electrical and Computer Engineering and the Coordinated Science Laboratory at the University of Illinois at Urbana-Champaign, Champaign IL, 61820 USA.
        {\tt\footnotesize \{andrems2,krdc\}@illinois.edu}}%
\thanks{Digital Object Identifier (DOI): 10.1109/LRA.2025.3563819}
}

\markboth{IEEE Robotics and Automation Letters. Preprint Version. Accepted April, 2025}
{Schreiber \MakeLowercase{\textit{et al.}}: Human-in-the-Loop Understanding for Fast Traversability Estimation}

\maketitle

\begin{abstract}
The increasing use of robots in unstructured environments necessitates the development of effective perception and navigation strategies to enable field robots to successfully perform their tasks. In particular, it is key for such robots to understand where in their environment they can and cannot travel---a task known as traversability estimation. However, existing geometric approaches to traversability estimation may fail to capture nuanced representations of traversability, whereas vision-based approaches typically either involve manually annotating a large number of images or require robot experience. In addition, existing methods can struggle to address domain shifts as they typically do not learn during deployment. To this end, we propose a human-in-the-loop (HiL) method for traversability estimation that prompts a human for annotations as-needed. Our method uses a foundation model to enable rapid learning on new annotations and to provide accurate predictions even when trained on a small number of quickly-provided HiL annotations. We extensively validate our method in simulation and on real-world data, and demonstrate that it can provide state-of-the-art traversability prediction performance. Code is available at: \url{https://github.com/andreschreiber/CHUNGUS}.
\end{abstract}


\IEEEpeerreviewmaketitle

\section{Introduction}
\IEEEPARstart{A}{s} technology advances, robots are increasingly being developed for use in harsh, hazardous environments. For example, field robots are being developed to automate tedious tasks in agriculture \cite{kayacan2018embedded} and to assist first responders in challenging areas like caves \cite{tranzatto2022cereberus}. With this expanded use of robots in challenging, unstructured environments, it is crucial that they can predict where they can and cannot travel. This challenge is known as traversability estimation and various methods have been proposed for tackling this problem \cite{papadakis2013terrain}.

Early methods for determining where a robot can travel adopted a geometric approach, where the environment was mapped using sensors such as LiDAR and representations like occupancy grids were used to plan an obstacle-free path \cite{moravec1985high, thrun2005probabilistic, siegwart2011introduction}. However, LiDAR sensors can be expensive, and purely geometric methods can struggle to capture the nuance seen in the environments encountered by field robots. Such challenges---along with the wide availability of cameras and advances in computer vision---have led to great interest in developing visual traversability estimation methods.

While a variety of approaches exist for using vision for navigation and traversability estimation, deep learning methods have become particularly popular in recent years \cite{maturana2017real, wellhausen2019where, wigness2019rugd, kahn2021badgr, gasparino2022wayfast, schmid2022self, castro2023how, schreiber2023weakly, schreiber2024wrizz, frey23fast, mattamala24wild, jung2024vstrong, triestvelociraptor, zhang2024traversability, gasparino2024wayfaster, ma2024imost, jiang2020rellis3d}. Nevertheless, these approaches present their own challenges. Segmentation methods \cite{maturana2017real, wigness2019rugd, jiang2020rellis3d} suffer from difficulty in defining an appropriate set of classes and involve a tedious labeling process \cite{schreiber2024wrizz, triestvelociraptor}. To remedy these issues, many recent approaches leverage self-supervision, which uses robot experience to learn a visual traversability model \cite{gasparino2022wayfast, schmid2022self, castro2023how,  frey23fast, mattamala24wild, jung2024vstrong, triestvelociraptor, zhang2024traversability, gasparino2024wayfaster, ma2024imost}. However, self-supervision introduces other issues, as it can produce noisy labels and needs additional sources of data \cite{schreiber2024wrizz}. Self-supervision also relies on robot experience (often via teleoperation), which makes it difficult to explicitly label non-traversable areas without risking damage to the robot or its environment. Furthermore, most learning-based methods do not learn traversability in an online fashion \cite{maturana2017real, wellhausen2019where, gasparino2022wayfast, schmid2022self, castro2023how, schreiber2023weakly, schreiber2024wrizz, jung2024vstrong, triestvelociraptor, gasparino2024wayfaster}, increasing these methods' susceptibility to performance degradation due to domain shift \cite{ma2024imost, yoon2024adaptive}.

\begin{figure}[tp]
  \vspace{2pt}
  \centering
  \includegraphics[width=3.4in]{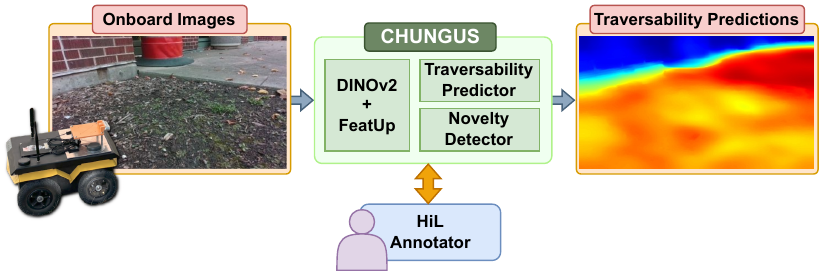}  \vspace{-10pt}
  \caption{Our proposed human-in-the-loop (HiL) traversability prediction method. A neural network built upon DINOv2 \cite{oquab2023dinov2} and FeatUp \cite{fu2024featup} predicts the traversability of each pixel in images collected by a robot. A novelty detector flags novel images and requests sparse annotations from a human when an unfamiliar image is detected. The sparse, easily-provided HiL annotations are used for online learning during robot deployment.}
  \vspace{-18pt}
  \label{fig:abstract}
\end{figure}

In contrast to existing self-supervised methods, we propose a new method, termed Continual Human Understanding for Navigational Guidance in Unstructured Settings (CHUNGUS). Our approach builds on prior work demonstrating the effectiveness of weakly-supervised learning using sparse, easily-collected manual traversability annotations \cite{schreiber2023weakly, schreiber2024wrizz}. We expand on such work and introduce an online traversability estimation method that prompts a human for a small number of sparse, relative annotations of traversability as-needed and rapidly retrains on the newly collected labels. An overview of our system is shown in Fig. \ref{fig:abstract}. We demonstrate that our method can quickly train a traversability neural network, and only requires a small number of labels from a human annotator that each take only seconds to collect. Our method can safely and explicitly label untraversable zones through its use of manual annotation and does not require a human demonstration or teleoperation phase, as is common in online learning methods~\cite{frey23fast,mattamala24wild,ma2024imost}. Instead, our framework efficiently prompts the human supervisor for sparse traversability annotations of images captured by the robot on an as-needed basis during robot deployment.

We summarize our contributions as follows:
\begin{enumerate}
    \item We propose a human-in-the-loop (HiL) visual traversability prediction method that uses sparse annotations to enable fast HiL labeling of images.
    \item We introduce an out-of-distribution image detection scheme to selectively request HiL annotations on an as-needed basis.
    \item We use vision foundation models \cite{oquab2023dinov2} to increase traversability prediction accuracy, which we combine with a feature upsampling method \cite{fu2024featup} to enable rapid from-scratch training of our traversability neural network during robot operation.
    \item We provide extensive evaluation of our method in a custom high visual fidelity simulator and on real-world data, and show that our approach provides state-of-the-art performance despite only using a small number of HiL labels that each take only seconds to provide.
    \item We demonstrate that our high visual fidelity simulator can produce sufficiently realistic images that enables models trained in simulation to produce accurate, high-quality traversability predictions on real-world data.
\end{enumerate}

\section{Related Works}
\subsection{Traditional Methods for Robot Navigation}
Early methods for robot navigation tended to adopt a geometric approach, where an environmental map is created and used to plan an obstacle-free path to a goal. Sensors like LiDAR can be used to detect objects, techniques like SLAM can be used to create a map and to localize the robot in the map~\cite{thrun2005probabilistic, siegwart2011introduction}, and representations like occupancy grids~\cite{moravec1985high} can be used for planning collision-free paths. However, LiDAR can be costly and purely geometric methods can struggle to capture important considerations. For example, in typical geometric approaches, traversable tall grass may appear as an untraversable obstacle~\cite{kahn2021badgr}, while loose surfaces (like sand or snow) may seem navigable from a purely geometric perspective but could cause the robot to slip or get stuck.

\subsection{Learning-Based Navigation and Traversability}
Seeking to address the limitations of non-learning-based geometric approaches, a variety of methods have been introduced that leverage machine learning and/or vision to provide more informed and nuanced traversability representations. For example, semantic segmentation models can be used to predict the semantic classes of each pixel in input images, and several datasets exist for segmentation in a field robotics setting \cite{wigness2019rugd, jiang2020rellis3d}. Maturana \textit{et al.} \cite{maturana2017real} demonstrate the use of such segmentation models for field robot navigation by fusing a geometric map with predictions from a semantic segmentation network to create a semantic map for navigation (where each semantic class can be assigned a navigation reward). However, segmentation approaches can be difficult to apply, as it is time-consuming to label data for semantic segmentation. In addition, it may be hard to define a suitable set of classes and it can be unclear how best to map these semantic classes into costs for navigation \cite{castro2023how, schreiber2024wrizz}.

The limitations of segmentation methods have led to significant interest in using self-supervision for traversability estimation and robot navigation. Kahn \textit{et al.} \cite{kahn2021badgr} present a self-supervised method (BADGR) that retroactively labels navigation events (e.g., bumpiness or collision) which are used to train an action-conditioned neural network for navigation.
Other approaches involve collecting proprioceptive measurements of terrain interactions that are used to 
generate labels for training a traversability prediction model \cite{wellhausen2019where,castro2023how}. Similarly, Gasparino \textit{et al.}~\cite{gasparino2022wayfast, gasparino2024wayfaster} explored using a known kinodynamic model to compute traction coefficients which are used as labels to train a traversability prediction neural~network.

Recent advancements in visual foundation models like DINO \cite{caron2021emerging}, Segment Anything Model (SAM) \cite{kirillov2023segment}, and DINOv2 \cite{oquab2023dinov2} have also been applied for traversability prediction. For example, Jung \textit{et al.} \cite{jung2024vstrong} introduce a self-supervised strategy for traversability estimation that uses SAM. Other work~\cite{triestvelociraptor, frey23fast, mattamala24wild} uses DINO or DINOv2 for computing high-quality visual features that can be used for self-supervised traversability prediction.

While self-supervised methods can remove the need for human annotation, such methods require robot experience and therefore cannot explicitly label untraversable areas without risking damage to the robot or its environment. Thus, these methods often fall into a positive and unlabeled (PU) learning setting \cite{seo2023learning, schreiber2024wrizz}. Moreover, these self-supervised methods often only perform learning in an offline manner \cite{kahn2021badgr, wellhausen2019where, gasparino2022wayfast, castro2023how, jung2024vstrong, triestvelociraptor, gasparino2024wayfaster}; therefore, they cannot adapt to changing conditions and struggle with issues like domain shift.
However, some more recent methods do incorporate online learning to combat this challenge. Wild Visual Navigation (WVN) \cite{frey23fast, mattamala24wild} quantifies traversability via discrepancy between commanded and achieved velocity and trains a neural network built on DINO \cite{caron2021emerging} to predict traversability in an online fashion. However, such an approach requires a human teleoperation phase and is not posed within the framework of continual learning. Other recent online traversability prediction methods do specifically adopt ideas from continual learning~\cite{ma2024imost,yoon2024adaptive} to mitigate issues like catastrophic forgetting~\cite{goodfellow2013empirical}; however, these methods either only use LiDAR (rather than vision)~\cite{yoon2024adaptive} or still require a human demonstration phase~\cite{ma2024imost}.

\subsection{Anomaly Detection for Field Robot Navigation}
Anomaly detection (AD) is a relevant area of study for our proposed method that involves detecting novel or out-of-distribution data samples. AD has been studied extensively for field robot navigation. For example, Wellhausen \textit{et al.}~\cite{wellhausen2020safe} develop an AD approach for navigation in unstructured environments. Similarly, Schmid \textit{et al.}~\cite{schmid2022self} introduce a self-supervised traversability prediction approach by training an autoencoder only on traversed areas of images (with reconstruction error being used as a proxy for traversability). WVN \cite{frey23fast,mattamala24wild}~also uses an AD approach based on reconstruction error to detect unfamiliar regions, but such approaches can be overly conservative as they correlate novelty or familiarity with traversability and may classify novel traversable regions as untraversable. In agricultural robotics, existing works \cite{ji2022proactive, schreiber2023attentional} propose proactive AD methods that utilize multi-modal inputs (camera and LiDAR) and a planned trajectory to predict the probability of future anomalies; however, these methods were not used directly for navigation due to the high computational cost of evaluating candidate paths.

\section{Method}
\begin{figure*}[tp]
  \centering
  \includegraphics[width=6.8in]{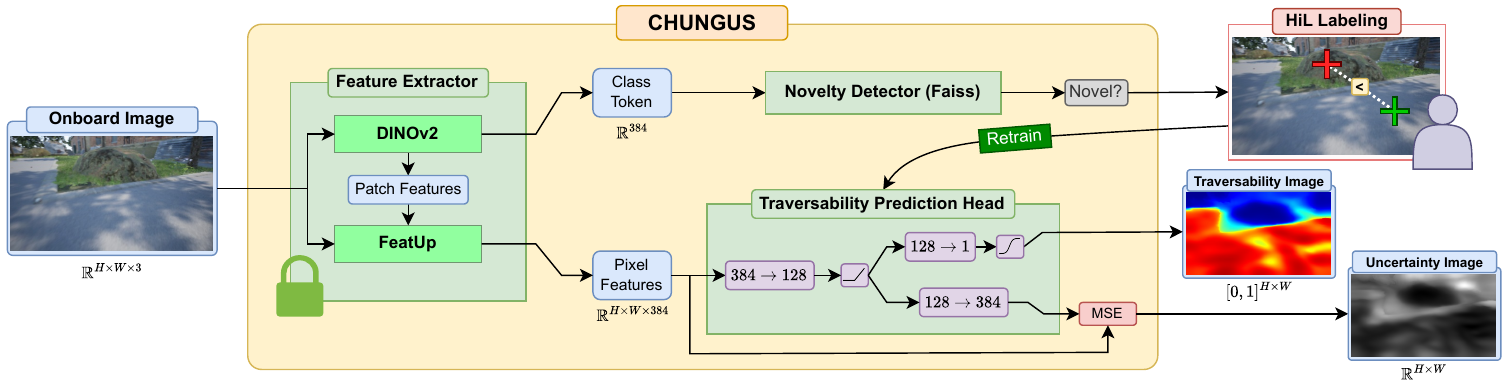}
  \vspace{-9pt}
  \caption{Overview of our proposed human-in-the-loop traversability prediction method. RGB camera images are passed through a DINOv2~\cite{oquab2023dinov2} feature extractor and FeatUp~\cite{fu2024featup} feature upsampler. The traversability prediction head takes upsampled image features as input and produces a traversability prediction (between 0 and 1) and an uncertainty score for each pixel. The class token from DINOv2 is fed to a novelty detector that detects novel images and requests labels for novel images from the HiL. The traversability prediction head is rapidly retrained when new HiL annotations are provided.}
  \label{fig:schematic}
  \vspace{-16pt}
\end{figure*}

We propose an online HiL traversability prediction system, CHUNGUS, that can learn during robot deployment by rapidly querying a human annotator for additional annotations when needed. As compared to existing vision-based approaches to traversability estimation, our proposed method does not require robot experience or teleoperation, only requires sparse labels (which take only seconds to collect) for training, and can perform online learning during robot deployment. Our method involves three key components: (1) a traversability prediction neural network based on DINOv2~\cite{oquab2023dinov2} and FeatUp~\cite{fu2024featup}, (2) an image anomaly/novelty detection scheme to determine when to ask the human for new annotations and retrain, and (3) a HiL annotation system. A schematic of our proposed system is shown in Fig. \ref{fig:schematic}.

\subsection{Network Architecture}\label{sec:architecture}
Similar to recent works in traversability estimation~\cite{frey23fast, mattamala24wild, jung2024vstrong, triestvelociraptor}, we utilize a visual foundation model. Specifically, we opt to use the ViT-Small variant of DINOv2 \cite{oquab2023dinov2}, which extracts rich visual features.
However, DINOv2 produces a single output feature for each $14 \times 14$ patch in the original image. Thus, naively using the DINOv2 patch features would only provide coarse predictions by producing outputs at a $14\times$ smaller resolution than the original input image. To address this issue, we apply state-of-the-art work in feature upsampling (FeatUp) \cite{fu2024featup} to upsample the features produced by DINOv2, and we then interpolate the upsampled features back to the original input image resolution to yield a 384-dimensional DINOv2 feature for each pixel in the input image. The DINOv2+FeatUp feature extractor can be viewed as a function that maps RGB images of resolution $H \times W$ to 384-D features of the same resolution: $f_\text{extractor}: \mathbb{R}^{H \times W \times 3} \mapsto \mathbb{R}^{H \times W \times 384}$.
Compared to upsampling the features using techniques like bilinear interpolation, the use of FeatUp \cite{fu2024featup} enables the upsampled features to retain critical details like edges, greatly improving prediction sharpness.

To produce traversability predictions from the upsampled DINOv2 features, a simple multi-layer perceptron (MLP) traversability prediction head network is used (where the MLP can be efficiently implemented via convolutional layers with a stride of 1 and a kernel size of 1). The MLP network contains a layer with a 384-dimensional input, a 128-dimensional output, and ReLU activation. This layer is then followed by a second layer that has 128-dimensional input and a 1-dimensional output with sigmoid activation, producing a traversability prediction $p^t \in [0,1]$. In addition, to quantify prediction uncertainty, we include a branch that takes as input the 128-dimensional output (after ReLU activation) from the first MLP layer and that produces a 384-dimensional output feature. We then compute the mean squared error (MSE) of this output 384-dimensional feature ($\mathbf{p}^r \in \mathbb{R}^{384}$) and the original DINOv2 image feature. We use this reconstruction MSE to quantify the novelty of each pixel in an input image (higher MSE indicates greater novelty).

To train our model, we use a similar strategy to Schreiber \textit{et al.} \cite{schreiber2024wrizz}, adopting the $\mathcal{L}_{\text{RIZZ}}$ loss and training on sparse, pairwise relative traversability labels. To train the reconstruction branch for a pixel, we minimize the MSE of the reconstructed upsampled DINOv2 feature and the original upsampled DINOv2 feature.

For training, we are given a pair of pixels with ordinal label $t \in \{-1,0,1\}$ ($t=-1$ indicates the first pixel in the pair is more traversable, $t=0$ means the pixels are equally traversable, and $t=1$ means the second pixel is more traversable), and we denote the upsampled DINOv2 features for the two pixels in the pair as $\mathbf{x}_a \in \mathbb{R}^{384}$ and $\mathbf{x}_b \in \mathbb{R}^{384}$ (respectively). We denote the predictions produced by the MLP traversability prediction head at the two pixels in the pair as $\mathbf{p}_a$ and $\mathbf{p}_b$ (respectively), where $\mathbf{p}^{r}_{a} \in \mathbb{R}^{384}$ and $\mathbf{p}^{r}_{b} \in \mathbb{R}^{384}$ are the reconstructed 384-dimensional DINOv2 features produced by the MLP head, and $p^{t}_{a} \in [0,1]$ and $p^{t}_{b} \in [0,1]$ are the traversability predictions produced by the MLP head. The loss used for training our model is given by:
\vspace{-6pt}
\begin{multline}
    \mathcal{L}_{\text{CHUNGUS}}(\mathbf{x}_a,\mathbf{p}_a,\mathbf{x}_b,\mathbf{p}_b,t) = \\ \mathcal{L}_{\text{RIZZ}}(p^{t}_a,p^{t}_b,t) + \gamma[\mathcal{L}_{\text{MSE}}(\mathbf{x}_a, \mathbf{p}^{r}_a) + \mathcal{L}_{\text{MSE}}(\mathbf{x}_b, \mathbf{p}^{r}_b)]
\end{multline}\label{eq_loss}
where $\mathcal{L}_{\text{MSE}}$ is a mean-squared error loss and $\gamma$ is a weighting term for the reconstruction loss. For all our experiments in this letter, we use a value of $\gamma=0.1$.

The DINOv2+FeatUp feature extractor module uses pretrained DINOv2 weights which remain frozen during traversability predictor training. Only the weights of the MLP head are updated during training. Of particular note, the MLP only needs to be trained on the sparse set of pairwise traversability annotations and the corresponding DINOv2 features at the annotated pixels, which---when combined with our sparse labeling strategy---enables our network to be trained from scratch extremely quickly (only a couple of seconds on a GPU using a dataset with thousands of images). The rapid training time of our method, paired with the relatively low storage cost of the sparse annotation labels and their corresponding 384-dimensional DINOv2 embeddings, allows us to circumvent catastrophic forgetting~\cite{goodfellow2013empirical} by simply caching all training data and retraining from scratch on all the cached data when new annotations are added. Despite using a relatively large foundation model, the inference speed of our network on an RTX 3080 GPU is $41 \ \text{ms}$, allowing for real-time prediction.

\subsection{Novelty Detection}\label{sec:novelty}
To determine when to prompt the human-in-the-loop annotator for additional annotations, we develop a novelty detection scheme. We base our scheme on the assumption that the network needs an annotation if the current image retrieved by the onboard camera is dissimilar to images that have been seen before. To quickly quantify novelty at the image level, we perform similarity search on the 384-dimensional class token produced by DINOv2. We utilize the Faiss \cite{douze2024faiss} library to quickly perform similarity search using an $L_2$ distance metric.

The class token for each image that our traversability prediction head is trained on is recorded in a Faiss index. To quantify the novelty of a new image from the robot's camera, the image's class token is queried against the Faiss index, and the distance (denoted $d_\text{new}$) between the nearest class token in the Faiss index to the class token of the new image is computed. If $d_\text{new} \geq \tau_{\text{novelty}}$ (where $\tau_{\text{novelty}}$ is a novelty threshold), the new image is classifed as novel and the human-in-the-loop will be asked to annotate the image.

For a Faiss index containing more than one class token, $\tau_{\text{novelty}}$ is computed from the distances of each class token in the index to its closest neighbor. If we denote the mean and standard deviation of the distances of each class token in the index to its nearest neighbor as $\mu_\text{class}$ and $\sigma_\text{class}$, respectively, the novelty threshold can be calculated as: $\tau_{\text{novelty}} = \mu_\text{class} + \alpha \sigma_\text{class}$, where $\alpha$ is a hyperparameter dictating how aggressively our system will ask for annotations from a human.

\subsection{HiL Annotation}\label{sec:hilannotation}
When the novelty detector detects an unfamiliar image, the image is sent to a human for annotation. The annotation procedure involves providing relative annotations for pixel pairs, where one pixel is labeled as more, less, or equally traversable  compared to the other pixel. For each image detected as novel, an annotation is provided for a pair of pixels where both pixels belong to the novel image (an intra-image label). A cross-image pair label is also provided, where one pixel in the pair belongs to the novel image and the other pixel belongs to a previously-labeled image \cite{schreiber2024wrizz}. The labeling interface displays the image to label, and the pair of pixels being labeled is indicated with colored crosshairs. The colors of the crosshairs specify the current label (the label is initialized using the current traversability network predictions), and the annotator can cycle through the label types and confirm the annotation using the keyboard. A video of the labeling procedure is provided in the supplemental material. The sparse annotation strategy enables rapid collection of annotations from the human-in-the-loop, taking only seconds for the annotator to label a new image.

The pixel locations for the intra-image label are selected randomly, and if the set of images that have been already labeled is non-empty, the other image to use for the cross-image label is selected randomly (cross-image labeling is skipped if the set of images that have been already labeled is empty). For cross-image labeling, the annotation pixel location for the new image is selected as the pixel having the greatest traversability prediction head reconstruction error on the upsampled DINOv2 features for the new image, while pixel location for the existing image (the randomly chosen image in our dataset of images that were already labeled) is the pixel in that image with the lowest reconstruction error. The selection of pixel locations for cross-image labeling is informed by the idea that we would like to label the most unfamiliar location in the new image by comparing it to something familiar. We use the altered strategy for cross-image labeling as we found that it provided improved performance compared to a random location selection strategy.

After the novel image is labeled by the human annotator, the traversability prediction neural network (described in Sec. \ref{sec:architecture}) is retrained from scratch on the new dataset that consists of all previously annotated images as well as the newly annotated image (we retrain the network from scratch on all annotations to avoid the issue of catastrophic forgetting). In addition, the novelty threshold is recomputed on this new data using the method in Sec. \ref{sec:novelty}. As traversing unlabeled novel areas may pose a risk to the robot, we adopt a conservative approach where the robot is commanded to stop while an image is being annotated and control resumes once the annotation and retraining is complete.

\subsection{Control Strategy}\label{sec:controller}
We seek to use our traversability predictions for robot navigation. We utilize a multi-modal elevation mapping system \cite{miki2022elevation,erni2023mem} to create a 2.5D elevation map of the environment, and the traversability predictions from our traversability neural network are fused into the elevation map as a custom elevation map layer. The fused traversability predictions then form a birds-eye-view (BEV) traversability map used for navigation.

Once the BEV traversability map is constructed, we follow existing work \cite{gasparino2022wayfast, gasparino2024wayfaster, schreiber2024wrizz} and utilize a non-linear model predictive controller for navigation. This controller uses the kinodynamic model (unicycle model) described by Gasparino \textit{et al.} \cite{gasparino2022wayfast} to simulate $N$ randomly sampled candidate control sequences (where the traversability coefficient $\mu$ for each state is set to the value corresponding to that state's location in the BEV traversability map and $\nu=1$). An optimization problem is constructed which seeks to minimize a cost that penalizes terminal state error, high control effort, and states being in low traversability zones. The optimization problem is solved using~the sampling-based Model Predictive Path Integral (MPPI) method \cite{williams2017information}.

\subsection{High Visual Fidelity Simulator}\label{sec:simulator}
\begin{figure}[tp]
  \centering
  \includegraphics[width=3.4in]{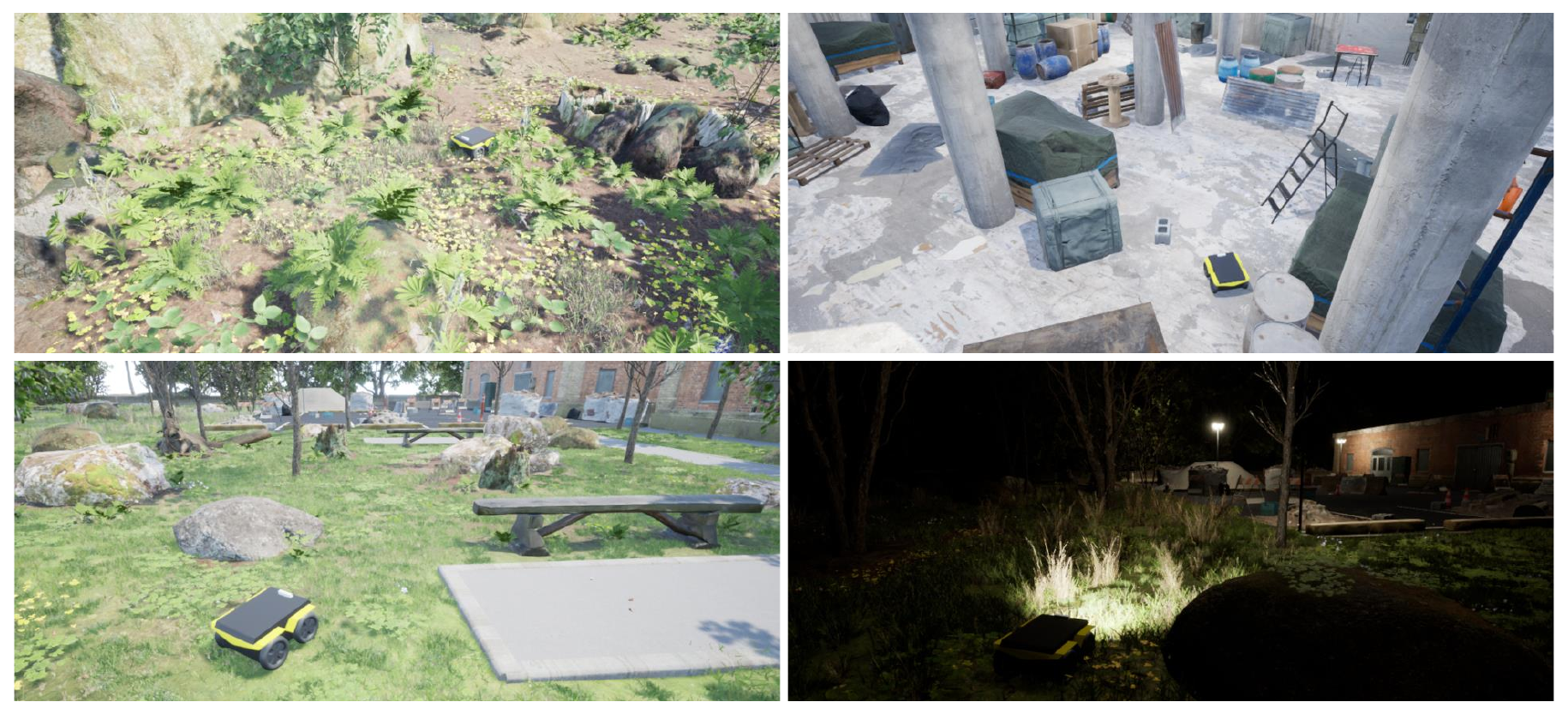}
  \vspace{-10pt}
  \caption{Images from our simulator, showing the forest (top left), warehouse (top right), lot (bottom left), and dark lot (bottom right) environments.}
  \label{fig:simulator}
  \vspace{-14pt}
\end{figure}

For evaluating our method, we construct a custom high visual fidelity simulator in Unreal Engine 4 (version 4.26). This simulator uses high-quality photo-scanned assets to provide photorealistic visuals. We simulate a small differential-drive robotic vehicle (based on a Clearpath Jackal) with various sensors (e.g., cameras) using the built-in physics system. A robot operating system (ROS) integration plugin \cite{mania2019framework} enables communication between the simulator and ROS 1 (Noetic), allowing for transmission of sensor data from the simulator to ROS nodes and to allow for control of the simulated robot by publishing twist commands in ROS. We implement several photorealistic environments in our simulator: a forest (``forest''), a cluttered warehouse (``warehouse''), and a parking lot with a wooded park area (``lot''). As a fourth environment, we use the same layout as ``lot'' but change the lighting to a dark (non-moonlit) night where only small portions of the map are lit with a flashlight and street lights (``dark lot''). Images taken from our simulator are shown in Fig. \ref{fig:simulator}.

\section{Experimental Results}
\subsection{Navigation Results}\label{sec:results_navigation}
We compare our method (CHUNGUS) with various baselines across navigation tasks in our four simulated environments. The environments feature challenging and varied conditions likely to be seen by field robots, including forests, tall grass, park-like areas, parking lots, cluttered buildings, and low-light conditions. The navigation tasks were as follows:
\begin{enumerate}
    \item[1)] \textit{Forest}: an approximately $50 \text{m}$ navigation task in the ``forest'' environment, where the robot must navigate to reach three waypoints.
    \item[2)] \textit{Warehouse}: an approximately $45 \text{m}$ navigation task in the ``warehouse'' environment, where the robot must navigate to reach two waypoints.
    \item[3)] \textit{Lot}: an approximately $50 \text{m}$ navigation task in the ``lot'' environment, where the robot must navigate to a goal (the goal is the only waypoint).
    \item[4.a)] \textit{Dark Lot}: the same task as \textit{Lot}, but in the low-light conditions of the ``dark lot'' environment.
    \item[4.b)] \textit{Dark Lot*}: an approximately $45 \text{m}$ navigation task in the ``dark lot'' environment, where the robot must navigate to a goal point (the goal is the only waypoint); the start and goal locations differ from that of \textit{Lot}.
\end{enumerate}

\begin{table*}[tp]
\centering
\caption{
Navigation Results in Simulation for Various Perception Strategies
}
\vspace{-6pt}
\label{tab:navigation_results}
\noindent
\resizebox{\textwidth}{!}{%
\begin{tabular}{c c c c c c c c c c c c}
\toprule
\multirow{3}{*}{Method}  & \multicolumn{2}{c}{Forest} & \multicolumn{2}{c}{Warehouse} &\multicolumn{2}{c}{Lot} & \multicolumn{2}{c}{Dark Lot} & \multicolumn{2}{c}{Dark Lot*}\\
\cmidrule(ll){2-3}\cmidrule(ll){4-5}\cmidrule(ll){6-7}\cmidrule(ll){8-9}\cmidrule(ll){10-11}

{} &
\makecell{Success\\Rate (\%)} & \makecell{Time ($\text{s}$)} &
\makecell{Success\\Rate (\%)} & \makecell{Time ($\text{s}$)} &
\makecell{Success\\Rate (\%)} & \makecell{Time ($\text{s}$)} &
\makecell{Success\\Rate (\%)} & \makecell{Time ($\text{s}$)} &
\makecell{Success\\Rate (\%)} & \makecell{Time ($\text{s}$)} &
\\
\midrule

\multicolumn{1}{l}{Blind} & $0$ & --- & $0$ & --- & $0$ & ---  & $0$ & --- & $0$ & --- \\
\multicolumn{1}{l}{LiDAR} & $0$ & --- & $60$ & $53.17$ & $10$ & $69.45$ & $10$ & $60.44$ & $90$ & $54.04$ \\
\multicolumn{1}{l}{WVN} & $10$ & $75.50$ & $40$ & $53.97$ & $30$ & $64.27$ & $10$ & $74.70$ & $20$ & $55.37$ \\
\multicolumn{1}{l}{WVN (5 min)} & $80$ & $71.62$ & $80$ & $58.13$ & $100$ & $66.32$ & $90$ & $65.19$ & $100$ & $57.31$\\
\midrule
\multicolumn{1}{l}{CHUNGUS ($\alpha=1$)} & $90$ & $130.46$ $(53.61)$ & $100$ & $131.75$ $(68.99)$ & $90$ & $132.14$ $(63.24)$ & $90$ & $131.27$ $(60.07)$ & $100$ & $115.06$ $(51.94)$ \\
\multicolumn{1}{l}{CHUNGUS ($\alpha=2$)} & $70$ & $94.30$ $(22.70)$ & $90$ & $85.76$ $(30.66)$ & $80$ & $86.09$ $(18.20)$ & $70$ & $93.34$ $(22.36)$ & $90$ & $86.05$ $(28.48)$ \\
\multicolumn{1}{l}{Big CHUNGUS} & $100$ & $68.40$ & $100$ & $53.98$ & $100$ & $64.05$ & $100$ & $62.60$ & $100$ & $56.56$ \\
\bottomrule
\end{tabular}
}
\vspace{-17pt}
\end{table*}

We evaluate the following methods to conduct a rigorous evaluation of our approach:
\begin{enumerate}
    \item \textit{Blind}: a naive blind pursuit method that assumes everywhere in the environment is traversable.
    \item \textit{LiDAR}: an obstacle-avoiding approach using an occupancy gridmap representation \cite{moravec1985high} created using a 2D LiDAR (with a $270 \degree$ field of view).
    \item \textit{WVN}: a state-of-the-art learning-based navigation approach \cite{mattamala24wild} trained as the robot navigates to its goal.
    \item \textit{WVN (5 min)}: the same model as \textit{WVN} where the weights are initialized by approximately 5 minutes of teleoperation in the environment under test.
    \item \textit{CHUNGUS ($\alpha=1$)}: our HiL method using a novelty detection hyperparameter value of $\alpha=1$.
    \item \textit{CHUNGUS ($\alpha=2$)}: our HiL method using $\alpha=2$.
    \item \textit{Big CHUNGUS}: a variant of our model using the newly proposed network architecture but trained on a dataset of 4858 images collected across all environments that are labeled as described by Schreiber \textit{et al.} \cite{schreiber2024wrizz}.
\end{enumerate}
The same controller is used for each method, such that the only difference between the methods is the perception and traversability prediction strategy. The simulator is run on a desktop PC with an RTX 2070 Super GPU, which is connected (via Ethernet) to a laptop with an RTX 4070 mobile GPU that runs all perception (including any online training), mapping, and control algorithms. The mapping system runs at $10 \ \text{Hz}$, while the controller runs at $10 \ \text{Hz}$ and samples $N=1024$ trajectories. Big CHUNGUS was trained on 4858 images prior to each navigation trial, which takes approximately 2 seconds.
The HiL CHUNGUS variants, CHUNGUS ($\alpha=1$) and CHUNGUS ($\alpha=2$), start each run with an empty training dataset and are only trained on HiL annotations collected during the current navigation trial.
For all experiments with CHUNGUS, camera images have a resolution of $H \times W = 240 \times 424$, which are resized to $224 \times 224$ for neural network inference. The neural network produces $224 \times 224$ resolution outputs that are then resized back to the original resolution of $240 \times 424$ to produce the final traversability predictions.

The results for our navigation experiments are presented in Table \ref{tab:navigation_results}, where we perform ten navigation trials per environment for each method and we measure the number of successful navigation trials for each method. We consider a run successful if all waypoints (including the goal) were reached. We also provide the median elapsed time to complete the navigation trial, measured over runs where navigation was successful. For HiL CHUNGUS, the median elapsed time spent for labeling for successful runs is in parentheses. Higher navigation success rate and lower elapsed time are better. Videos from these experiments are provided in the supplemental material.

\pagebreak 
The results in Table \ref{tab:navigation_results} demonstrate that Big CHUNGUS performs best, successfully navigating in all attempts. The blind strategy failed in every navigation attempt by crashing into obstacles. In addition, the LiDAR strategy often performs poorly, as it frequently struggled to detect crucial obstacles (e.g., parking blocks or rocks) below the measurement plane of the LiDAR and suffered from noise when traversing bumpy surfaces.
The results for \textit{Lot} and \textit{Dark Lot} also demonstrate that the methods show generally similar success rates on the same scenario in day and night conditions (with a minor drop in success rate for some of the methods at night).

The results in Table \ref{tab:navigation_results} indicate that WVN \cite{mattamala24wild} struggles to perform well (achieving success rates of less than $50\%$). By comparison, our HiL CHUNGUS models achieve significantly higher navigation success rates despite also only being trained while the robot is deployed. The improved success rate is due to our method being able to determine when the robot encounters novel data and requesting for additional labels in such cases (pausing navigation until annotations are provided), rather than needing the robot to experience conditions to label them and requiring a teleoperation phase to be trained effectively. The performance of WVN significantly increases when given a 5 minute teleoperation phase prior to the navigation task; nonetheless, we see that WVN with 5 minutes of teleoperation still performs worse than Big CHUNGUS and performs similarly to the HiL variants of CHUNGUS (despite HiL CHUNGUS not needing a teleoperation phase and being trained only during deployment). We also see that using a smaller $\alpha$ for HiL CHUNGUS leads to greater navigation success rate (by asking for more annotations) at the expense of slightly longer navigation times. However, even with $\alpha=2$, CHUNGUS provides relatively high navigation success rates.

The timing results in Table \ref{tab:navigation_results} demonstrate that all non-HiL methods take similar times to navigate to the goal. The HiL models take longer to complete the task due to navigation pauses during HiL annotation. However, the added labeling time is relatively small, typically only adding $30$ seconds ($\alpha=2$) to a minute ($\alpha=1$) to the navigation time. This added time due to HiL labeling is significantly less than the human teleoperation time given to the WVN (5 min) variant, despite HiL CHUNGUS performing similarly to WVN (5 min). During our navigation trials, it took only ${\sim}0.1 \ \text{s}$ for each retraining of HiL CHUNGUS on the small number of HiL annotated images collected during deployment, and for successful trials the total time spent training HiL CHUNGUS for a given run is ${\sim} 1 \ \text{s}$ (for $\alpha=2$) or ${\sim} 2 \ \text{s}$ (for $\alpha=1$).
Finally, CHUNGUS provided an inference rate of 7.7 Hz for all variants,
whereas WVN had an inference rate of 5.7 Hz.

\begin{table}
\caption{Anomaly Detection Performance}
\vspace{-12pt}
\label{tab:anomaly_detect}
\begin{center}
\begin{tabular}{@{} l *{4}{r} @{}}
\toprule
{Distance Metric} & \multicolumn{1}{c}{$\text{Accuracy}$} & \multicolumn{1}{c}{$\text{F1-score}$} & \multicolumn{1}{c}{$\text{ROC-AUC}$} & \multicolumn{1}{c}{$\text{PR-AUC}$} \\
\midrule
{$L_2$} & \multicolumn{1}{c}{0.917} & \multicolumn{1}{c}{0.937} & \multicolumn{1}{c}{0.989} & \multicolumn{1}{c}{0.996} \\
{$\textit{Cosine}$} & \multicolumn{1}{c}{0.912} & \multicolumn{1}{c}{0.933} & \multicolumn{1}{c}{0.987} & \multicolumn{1}{c}{0.995}\\
\bottomrule
\end{tabular}
\end{center}
\vspace{-18pt}
\end{table}

\subsection{Novelty Detection Results}\label{sec:results_novelty}
We additionally analyze the performance of our novelty detection strategy. We use a dataset of $4858$ images across the four simulated environments (divided into a training set of $4000$ images and validation set of $858$ images). The training and validation datasets are each split into four subsets, corresponding to the images from each environment (forest, warehouse, lot, and dark lot). We create our novelty predictor by using the training set images corresponding to a given environment, and we evaluate accuracy, F1-score, area under the receiver operating characteristic curve (ROC-AUC), and area under the precision-recall curve (PR-AUC) on all validation set images. We consider images from the same environment as used to train the index as normal samples and the images from different environments as anomalous. We evaluate on each of the four environments and provide results in Table \ref{tab:anomaly_detect}. For threshold-dependent metrics (accuracy and F1-score), we use an anomaly threshold calculated as described in Sec. \ref{sec:novelty} with $\alpha=2$ (the threshold is computed separately for each of the four environments using data from the training dataset). The metrics are calculated individually for each of the four environments, and the results in Table \ref{tab:anomaly_detect} are the average values for each metric across the four environments. For each metric in Table \ref{tab:anomaly_detect}, higher is better. We show results using $L_2$ as the distance in our Faiss index and using cosine distance ($\textit{Cosine}$).

The results in Table \ref{tab:anomaly_detect} show that our novelty detection strategy can detect novelty with high accuracy. Moreover, such results show that using $L_2$ distance outperforms cosine distance, but the two methods perform very similarly overall.

\subsection{Ablation Study of Navigation Performance}
To further evaluate our design decisions, we conduct an extensive ablation study of our HiL annotation scheme and present the results in Table \ref{tab:navigation_ablation}, where we show navigation success rate, median number of annotated images for successful runs, and median time needed for navigation for successful runs. For this study, we perform experiments on the \textit{Lot} and \textit{Warehouse} navigation tasks described in Sec. \ref{sec:results_navigation}, where each method is run ten times in each environment. We investigate the use of Faiss for novelty detection as described in Sec. \ref{sec:novelty} using an $L_2$ distance metric with various values of the novelty hyperparameter $\alpha$.

In addition to analyzing the value of $\alpha$, we analyze several approaches to selecting the pixel locations for the HiL annotations: $\textit{Random}$, which selects locations for intra- and cross-image labels randomly; \textit{Smart}, which uses the label selection strategy described in Sec. \ref{sec:hilannotation}; \textit{Smart-Both}, which selects the cross-image labels as in Sec. \ref{sec:hilannotation} and selects the intra-image annotation pixels such that one pixel corresponds to the pixel with the greatest reconstruction error and the other corresponds to the pixel with the smallest reconstruction error; and \textit{Manual}, where the human annotator manually selects the pixel locations for all annotations.

\begin{table}[tp]
\caption{Navigation Ablation Study}
\label{tab:navigation_ablation}
\vspace{-12pt}
\begin{center}
\begin{tabular}{@{} l *{5}{r} @{}}
\toprule
\multicolumn{1}{c}{Map} &
\multicolumn{1}{c}{\makecell{Labeling \\ Method}} & 
\multicolumn{1}{c}{\makecell{$\alpha$}} & \multicolumn{1}{c}{\makecell{Success \\ Rate ($\%$)}} & \multicolumn{1}{c}{\makecell{Num. Images \\ Annotated}} &  \multicolumn{1}{c}{\makecell{Time (s)}} \\
\midrule

Lot & \multicolumn{1}{c}{Smart} & \multicolumn{1}{c}{$1$} & \multicolumn{1}{c}{$90$} & \multicolumn{1}{c}{$20$} & \multicolumn{1}{c}{$132.14$} \\
& \multicolumn{1}{c}{Random} & \multicolumn{1}{c}{$2$} & \multicolumn{1}{c}{$70$} & \multicolumn{1}{c}{$8$} & \multicolumn{1}{c}{$87.92$} \\
& \multicolumn{1}{c}{Smart} & \multicolumn{1}{c}{$2$} & \multicolumn{1}{c}{$80$} & \multicolumn{1}{c}{$7.5$} & \multicolumn{1}{c}{$86.09$} \\
& \multicolumn{1}{c}{Smart-Both} & \multicolumn{1}{c}{$2$} & \multicolumn{1}{c}{$50$} & \multicolumn{1}{c}{$7$} & \multicolumn{1}{c}{$88.06$} \\
& \multicolumn{1}{c}{Manual} & \multicolumn{1}{c}{$2$} & \multicolumn{1}{c}{$80$} & \multicolumn{1}{c}{$8.5$} & \multicolumn{1}{c}{$111.94$} \\
& \multicolumn{1}{c}{Smart} & \multicolumn{1}{c}{$3$} & \multicolumn{1}{c}{$60$} & \multicolumn{1}{c}{$5.5$} & \multicolumn{1}{c}{$77.71$} \\
\midrule
Warehouse & \multicolumn{1}{c}{Smart} & \multicolumn{1}{c}{$1$} & \multicolumn{1}{c}{$100$} & \multicolumn{1}{c}{$26$} & \multicolumn{1}{c}{$131.75$} \\
& \multicolumn{1}{c}{Random} & \multicolumn{1}{c}{$2$} & \multicolumn{1}{c}{$70$} & \multicolumn{1}{c}{$11$} & \multicolumn{1}{c}{$91.00$} \\
& \multicolumn{1}{c}{Smart} & \multicolumn{1}{c}{$2$} & \multicolumn{1}{c}{$90$} & \multicolumn{1}{c}{$9$} & \multicolumn{1}{c}{$85.76$} \\
& \multicolumn{1}{c}{Smart-Both} & \multicolumn{1}{c}{$2$} & \multicolumn{1}{c}{$70$} & \multicolumn{1}{c}{$10$} & \multicolumn{1}{c}{$86.72$} \\
& \multicolumn{1}{c}{Manual} & \multicolumn{1}{c}{$2$} & \multicolumn{1}{c}{$90$} & \multicolumn{1}{c}{$10$} & \multicolumn{1}{c}{$122.60$} \\
& \multicolumn{1}{c}{Smart} & \multicolumn{1}{c}{$3$} & \multicolumn{1}{c}{$80$} & \multicolumn{1}{c}{$7.5$} & \multicolumn{1}{c}{$77.37$} \\
\bottomrule
\end{tabular}
\end{center}
\vspace{-22pt}
\end{table}

The results in Table \ref{tab:navigation_ablation} show that HiL CHUNGUS requires a very small number of annotated images to enable successful navigation. Moreover, the results demonstrate that the smart labeling strategy used by CHUNGUS outperforms the random method and the smart-both method of labeling and performs the same as the manual method in terms of navigation success rate. All non-manual methods showed approximately the same time needed to provide each intra- or cross-image label with an average time of ${\sim}1.5 \ \text{s}$ needed to provide a single label, while the manual method took ${\sim}3.3 \ \text{s}$ to provide each label (since pixel locations also needed to be manually chosen). Finally, increasing $\alpha$ speeds up navigation at the expense of a slightly lower navigation success rate, as a larger value of $\alpha$ leads to a stricter criterion of novelty and fewer HiL annotations.

\subsection{Evaluation on Real-World Data}

As our method is developed and tested in a high visual fidelity simulator, we seek to investigate the generalization performance of models trained on simulator data when applied to real-world images. For these experiments, we use the Big CHUNGUS model described in Sec. \ref{sec:results_navigation}. We utilize two datasets: \textit{Sim} (with $4000$ training and $858$ validation images as described in Secs. \ref{sec:results_navigation} and \ref{sec:results_novelty}) and \textit{Real} (containing $7342$ images collected with a Clearpath Jackal and a RealSense D435i camera). The \textit{Real} dataset is collected in three distinct areas (a semi-urban campus, a park/grove, and a wooded area with tall grass) and is split into $4000$ training and $3342$ validation images. The datasets have the same number of training images to enable fair comparison.

We train two variants of our Big CHUNGUS model, with one trained solely on the sim training dataset and the other trained solely on the real training dataset. We measure Human Disagreement Rate (HDR) at various thresholds \cite{schreiber2024wrizz} for the sim model evaluated on the sim validation set and the real validation set, as well as the real model evaluated on the real validation set. The results are presented in Table \ref{tab:sim2real} (where lower HDR is better) and show that the model trained only on simulator data shows only a relatively modest increase in $\text{HDR}_{0.1}$ and $\text{HDR}_{0.25}$ and has a lower $\text{HDR}_{0.5}$.

\begin{table}
\caption{Sim2Real Transfer with Big CHUNGUS}
\label{tab:sim2real}
\vspace{-12pt}
\begin{center}
\begin{tabular}{@{} l *{4}{r} @{}}
\toprule
\makecell{Training Data} & \makecell{Evaluation Data} & \multicolumn{1}{c}{$\text{HDR}_{0.1}$} & \multicolumn{1}{c}{$\text{HDR}_{0.25}$} & \multicolumn{1}{c}{$\text{HDR}_{0.5}$} \\
\midrule
{Sim} & \multicolumn{1}{c}{Sim} & \multicolumn{1}{c}{$0.152$} & \multicolumn{1}{c}{$0.107$} & \multicolumn{1}{c}{$0.267$} \\
\midrule
{Sim} & \multicolumn{1}{c}{Real} & \multicolumn{1}{c}{$0.226$} & \multicolumn{1}{c}{$0.157$} & \multicolumn{1}{c}{$\mathbf{0.283}$} \\
{Real} & \multicolumn{1}{c}{Real} & \multicolumn{1}{c}{$\mathbf{0.159}$} & \multicolumn{1}{c}{$\mathbf{0.126}$} & \multicolumn{1}{c}{$0.287$} \\
\bottomrule
\end{tabular}
\end{center}
\vspace{-16pt}
\end{table}

\begin{figure}[tp]
  \centering
  \includegraphics[width=3.4in]{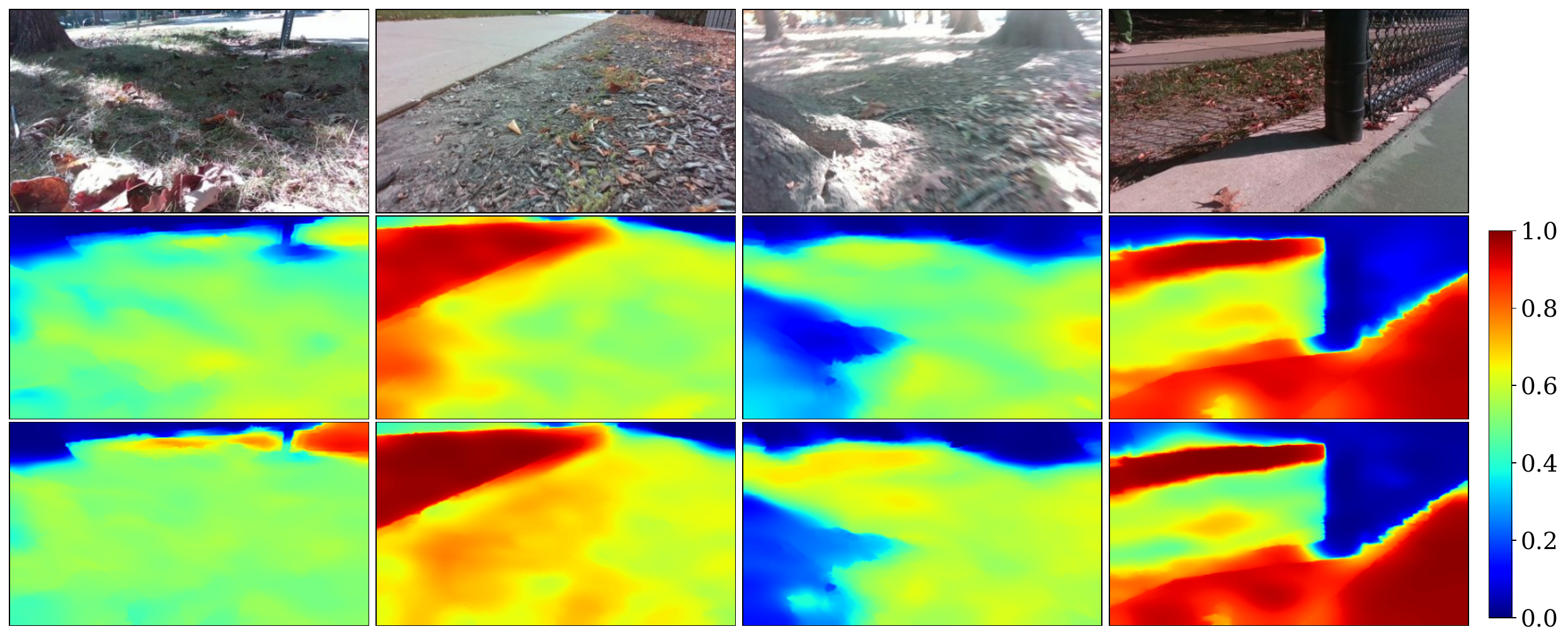}
  \vspace{-10pt}
  \caption{Traversability predictions using Big CHUNGUS, showing color images (top), predictions from a model trained only on data from our simulator (middle), and predictions from a model trained only on real-world data (bottom). The traversability values range from 0 (least traversable) to 1 (most traversable). Higher traversability is indicated by warmer colors.}
  \label{fig:sim2real}
  \vspace{-16pt}
\end{figure}

We further compare qualitative predictions using the sim-trained Big CHUNGUS and real-trained Big CHUNGUS in Fig. \ref{fig:sim2real}. These qualitative predictions demonstrate that the variant of Big CHUNGUS trained solely on simulator data can provide accurate, high-quality traversability predictions on real-world data (with striking qualitative similarity to the predictions from the model trained on real-world data). Videos showing predictions from Big CHUNGUS on real-world data are provided as supplemental material.

\section{Conclusion}
We introduce a novel human-in-the-loop traversability learning method that can successfully train a traversability prediction neural network during robot deployment using only a small number of labels quickly acquired from the human annotator. We develop a novelty detection scheme to detect novel environments and prompt an annotator for additional labels as-needed. We use foundation models~\cite{oquab2023dinov2} and image feature upsampling methods~\cite{fu2024featup} to enable high-quality traversability predictions and rapid online training. Our method is extensively evaluated in a photorealistic simulator, and our experiments show that our method outperforms a state-of-the-art online self-supervised traversability prediction method. Finally, we demonstrate that the high visual fidelity of the simulator enables effective Sim2Real transfer of our visual traversability prediction model. While our method does still show the limitation of needing a HiL labeler, the time needed for labeling is small and our method does not require teleoperation. Future work involves evaluating our approach on more vehicle platforms (in simulation and on physical hardware), as well as combining our method with self-supervision to further improve labeling efficiency.

\section*{Acknowledgment}
The authors gratefully acknowledge the Army Corps of Engineers Engineering Research and Development Center, Construction Engineering Research Laboratory for their review and approval of the final manuscript and financial support under contract award number W9132T23C0023.  We would also like to thank William R. Norris and Ahmet Soylemezoglu for feedback throughout the project and for providing access to robot hardware for collecting real-world data, Mateus Gasparino for his insights and for providing reference code for the controller, and Aamir Hasan for his valuable comments on early drafts.

\ifCLASSOPTIONcaptionsoff
  \newpage
\fi

\bibliographystyle{IEEEtran}
\bibliography{references}

\end{document}